\documentclass{article}

\usepackage{arxiv}

\usepackage[utf8]{inputenc} 
\usepackage[T1]{fontenc}    
\usepackage{hyperref}       
\usepackage{url}            
\usepackage{booktabs}       
\usepackage{amsfonts}       
\usepackage{nicefrac}       
\usepackage{microtype}      
\usepackage{lipsum}		
\usepackage{graphicx}
\usepackage{doi}

\usepackage{relsize}
\usepackage{multicol,multirow}

\title{An Ontology-Based Information Extraction System for Residential Land Use Suitability Analysis}

\author{ Munira Al-Ageili \\
	Department of Computer Science\\
	  University of Regina\\
      \texttt{munira.al-ageili@uregina.ca} \\
	\And
	\href{https://orcid.org/0000-0001-7381-1064}{\includegraphics[scale=0.06]{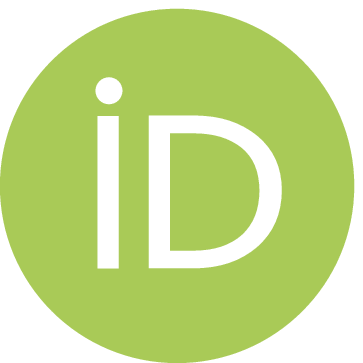}\hspace{1mm}Malek Mouhoub} \\
 	Department of Computer Science\\
	  University of Regina\\
	\texttt{mouhoubm@uregina.ca} \\
}

\renewcommand{\shorttitle}{\textit{arXiv} Template}

\hypersetup{
pdftitle={A template for the arxiv style},
pdfsubject={q-bio.NC, q-bio.QM},
pdfauthor={David S.~Hippocampus, Elias D.~Striatum},
pdfkeywords={First keyword, Second keyword, More},
}

\begin{document}
\maketitle

\begin{abstract}
  We propose an Ontology-Based Information Extraction (OBIE) system to automate the extraction of the criteria and values applied in Land Use Suitability Analysis (LUSA) from bylaw and regulation documents related to the geographic area of interest. The results obtained by our proposed LUSA OBIE system (land use suitability criteria and their values) are presented as an ontology populated with instances of the extracted criteria and property values. This latter output ontology is incorporated into a Multi-Criteria Decision Making (MCDM) model applied for constructing suitability maps for different kinds of land uses. The resulting maps may be the final desired product or can be incorporated into the cellular automata urban modeling and simulation for predicting future urban growth. A case study has been conducted where the output from LUSA OBIE is applied to help produce a suitability map for the City of Regina, Saskatchewan, to assist in the identification of suitable areas for residential development. A set of Saskatchewan bylaw and regulation documents were downloaded and input to the LUSA OBIE system. We accessed the extracted information using both the populated LUSA ontology and the set of annotated documents. In this regard, the LUSA OBIE system was effective in producing a final suitability map.
\end{abstract}


\section{Introduction}
Land use suitability analysis is used to assess the appropriateness of a specific area of land for a particular use \cite{ogunjobi2018spatio,sang2019intensity,malczewski2004gis}. In the context of Geographic Information Systems (GIS), land use suitability is determined through a systematic multi-factor analysis of the different aspects of the landscape. Model input, therefore, can include factors related to physical and environmental sustainability, as well as factors pertaining to economic and cultural impacts. The results of the analysis are usually presented on maps that rate areas from high to low suitability. The maps may be the desired end result or they may be used as one of the inputs to a simulation model, such as cellular automata, for representing and predicting the spatial dynamics of land use such as urban growth.
In this modeling approach, the most important task is specifying the criteria and values that are applied to assess the suitability of the land for a particular kind of use in this study, for residential development. Determining the factors and their criteria and values helps in determining the data sets needed to create the GIS layers to be included in the evaluation.\\

 Each jurisdiction has its own regulations, bylaws, or policies that are applied to assess land use suitability for that jurisdiction geographic area.  These regulations provide the criteria for the factors to be included in the multi-criteria analysis (see Table \ref{tab:addlabel}).   Bylaw and regulation documents available on the web are in natural language text and cannot be processed directly by machines to access and extract the information. Manually finding and extracting criteria and the specific values for the criteria can be a tedious and time-consuming task.  In some cases, there may be no precise values for criteria and the actual or real values require expert judgment.
For the purpose of automating the extraction of the criteria and values applied in land use suitability analysis (LUSA) \cite{parry2018gis}, we developed an ontology-based information extraction (OBIE) system to automatically extract the required information from bylaw and regulation documents related to the geographic area of interest.\\

The output of the LUSA OBIE system can be presented as an Geospatial ontology \cite{DBLP:conf/geos/BaglioniMRS07,muller2004textpresso,sun2019geospatial,Hathcoat2017,Bontcheva2004,Gruber:1993:TAP:173743.173747,Gruber:1995:TPD:219666.219701,Smith2003-SMIO-2} populated with instances of the extracted criteria and property values, as a set of semantically (with ontology knowledge)  annotated documents or the populated LUSA ontology can be exported to a database or a knowledge base or can be saved as an XML file. The user can retrieve the information from the ontology, view the annotated documents using an annotation editor, or query the database or knowledge base. The extracted criteria are then used to direct the process of obtaining the necessary data for the creation of the required land use suitability maps and also for determining the GIS operations that should be performed to create the GIS layers that represent the criteria.
The output from LUSA OBIE is applied in this paper to help produce a suitability map for the City of Regina, Saskatchewan to assist in the identification of suitable areas for residential development. A set of Saskatchewan bylaw and regulation documents were downloaded and input to the LUSA OBIE system. We accessed the extracted information (criteria and data property values) using both the populated LUSA ontology and the set of annotated documents.\\

The rest of the paper is organized as follows. The next section describes the spatial relations considered for this study. Our proposed method is then presented in Section 3. The case study for the city of Regina is described in Section 4. Finally concluding remarks are listed in Section 5. Note that this paper extends the previous work  we conducted in information extraction for residential land use suitability \cite{al2015ontology}.

\section{{Decision Criteria Based on Spatial Relations }}
Spatial relations between geographic objects are key elements of spatial modeling and spatial analysis. These relationships among objects in space result from their locations relative to each other. Geographic Information Systems (GIS) are often based upon spatial relations. A primary function of these systems is determining the relationships between objects in space.\\

Spatial relations are classified into topological, orientation, and distance relations \cite{Pullar88,Sharma96}. Topological relations describe the spatial relation between neighboring features such as adjacency, connectivity and containment. These relations are purely qualitative, and invariant under continuous transformations such as rotation and scaling. Direction relations help determine the orientation of a primary object relative to a reference object; they could be qualitative, such as ``{\it Regina is east of Moose Jaw}'', or metric, such as a direction specified in degrees from a reference direction. Distance relations specify the distance between two objects; they could be metric distance relations, such as ``{\it the distance separating a hazardous waste site and a residential subdivision should be at least 2 kilometers}'', or qualitative distance relations, such as near, close to, far, and very far; for example, ``{\it the house is close to the main road}''.\\

Criteria (factors and constraints) involved in a multi-criteria analysis describe some of the spatial characteristics of the area under consideration. For example, in a land use suitability analysis for residential development, these criteria may include topographic properties of the land, such as slope of the terrain, and accessibility to amenities and services such as parks, roads, and fire, police and ambulance services.
Spatial factors have a significant role in a land use suitability analysis process. Factors and constraints criteria are usually defined spatially in a way that depends on the GIS model (Raster or Vector) used to create them. Some criteria such as ''{\it distance to roads}''   are explicitly spatial and are usually created using GIS functionality. These criteria are represented as spatial data layers.
Land use suitability evaluation criteria may include factors and constraints related to the physical attributes of the land (e.g., topography, soil characteristics, potential flooding, subsidence and erosion, and servicing). In \cite{rinner2006spatial}, the authors
  distinguish between three classes of spatial relations applicable to multi-criteria decision-making: the location of the sites under consideration, their proximity to desirable or undesirable facilities or features, and the direction relative to certain facilities and the sites under consideration.\\

The spatial relations primarily applicable to land use suitability analysis criteria are the location of the sites (or subdivisions) under consideration and their proximity to desirable or undesirable facilities. Direction relations are also important in some cases, such as directions of aircraft takeoff and landing relative to the location of the residential subdivision. However, direction relations are not dealt with in the selected domain documents. In addition, some criteria describe the spatial characteristics of the land, such as soil type, soil conditions, and slope or aspect of the land. Examples related to land use suitability analysis for residential development are shown in Table \ref{tab:addlabel}\footnote{Saskatchewan subdivision guidelines. Retrieved from http://www.municipal.gov.sk.ca/Subdivision/Subdivision-Guide}.

\begin{table*}[htbp]
{\relsize{-2}
  \centering
  \caption{Criteria examples extracted from selected domain documents }

 \begin{center}
    \begin{tabular}{|l|l|l|}
   \hline
    \multicolumn{1}{|c}{\textbf{Spatial Relation/}} & \multicolumn{1}{|c}{\textbf{Criterion}} & \multicolumn{1}{|c|}{\textbf{Example}} \\
  \multicolumn{1}{|c}{\textbf{Characteristic}} & \multicolumn{1}{|c}{\textbf{}} & \multicolumn{1}{|c|}{\textbf{}} \\ \hline
    \multirow{3}[2]{*}{Location} & Permitted land use & Subdivision zoned as residential  \\
   \multirow{3}[2]{*}{(derived from topology)}          &       &  \\
          &       & Building site on or near a drop off \\
          & Topography &  \\
    \multirow{5}[2]{*}{Proximity (distance)} & Servicing   & Access to roads,   \\
     \multirow{5}[2]{*}{ } &  (access to desirable facilities) &  water or sewer connection \\
          &       &  \\
          & Setbacks  & Less than 457 meters from  \\
           &   (close to undesirable facility) & a landfill sewage\\
          &       &   treatment   plant;\\
          & Neighboring  land use   &  mining   facility industrial \\
          & (close to undesirable area) &  development  \\
    \multirow{6}[2]{*}{Land characteristics} & Soil type & Unsuitable soil type such as   \\
       \multirow{6}[2]{*}{ } &   &   Loose or swampy soil \\
          &       & Soils shifting, heaving or cracking \\
          & soil condition & Steeply sloping land \\
          &       & Polluted drainage onto the land  \\
          &       & from adjacent uses \\
          & Slope &  \\
          & Surface and sub-surface drainage & \\ \hline
    \end{tabular}%
    \end{center}
  \label{tab:addlabel}%
  }
\end{table*}%

\section{OBIE for Land Use Suitability Analysis Criteria (LUSA)}

We propose a framework for integrating an  OBIE  system and GIS-Based Multi-Criteria Decision Making approach (MCDM). The objective is to construct a land suitability map for residential development for the City of Regina.
We developed an OBIE system for automating the extraction of criteria and their values that are applied in land-use suitability analysis, from bylaws and regulations documents, related to the geographic area of interest. These criteria represent the biophysical, social and economic factors that may be used in the construction of land-use suitability maps that support the process of evaluating the suitability of a particular area of land for a particular kind of use.
The results obtained by LUSA OBIE (land use suitability criteria and their values) are then incorporated into the MCDM model applied for constructing a suitability map for residential development for the City of Regina.

\begin{figure}
 \centering
\includegraphics[width=9.5cm ]{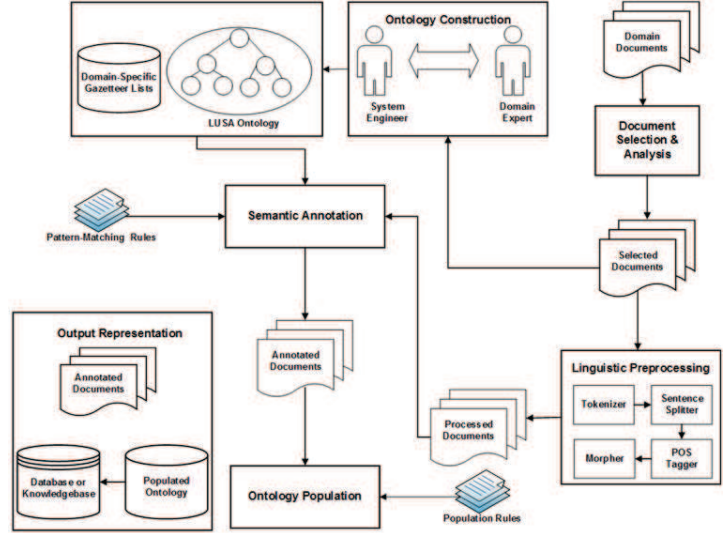}
\caption{LUSA OBIE Architecture}
\label{lusa}
\end{figure}

The LUSA OBIE system combines the use of ontology, domain-specific gazetteer lists, language processing tools and extraction rules based on regular expressions  \cite{cunningham2011text} to automatically add semantic annotation to domain documents (such as regulations and bylaws documents) and then extract the criteria to be applied in the process of land use suitability assessment. Figure \ref{lusa} illustrates the overall architecture and the following components of   LUSA OBIE.

\subsection{Documents Selection and Analysis.} A set of relevant bylaw documents in natural language, related to the geographic area of interest, is selected.  The selected documents are carefully examined to help identify domain concepts to be included in the domain ontology, as well as for enumerating domain-specific gazetteer lists.

One of the requirements of IE is that the type of content to be extracted must be predefined. Therefore, we need to identify and examine domain-specific documents in order to identify relevant concepts that should be included in the ontology that will guide the information extraction process. The overall or general domain of interest is land use suitability; the specific application domain dealt with in this work is land use suitability assessment in Saskatchewan. Documents relevant to the Province of Saskatchewan were searched for, examined for basic relevance, and ''“ where appropriate ''“ downloaded from the internet. These documents include information about the regulations and policies that are applied to assess land use suitability in the province. Selected documents pertaining to land use suitability in Saskatchewan were subsequently examined in detail to help provide the concepts to assist in constructing the ontology and to help provide the terminology to be used in gazetteer list construction and for building pattern extraction rules. The primary concepts that were automatically identified include: Topography, Soils, Surface and sub-surface drainage, Potential for flooding and other hazards, Easements (or Interests), Land use (uses) in the vicinity, Streets, lanes, traffic flow and public safety, Site design and orientation, the protection of fish and wildlife habitats, the protection of significant natural and historical features, Setbacks and Public Lands.

\subsection{Linguistic Preprocessing.}

 The selected domain documents are written in natural language, therefore it is difficult to directly process them and extract the information. The input text needs first to be structured to identify its essential lexical and syntactic constituents and make the knowledge accessible. We use a set of linguistic processing tools to process the text to obtain its various linguistic features.  Text preprocessing includes tokenization (i.e., splitting the text into tokens), sentence splitting (segmenting the text into sentences), shallow lexico-syntactic analysis (such as Part of Speech (POS) \cite{tian2015comparative} tagging and morphological analysis) and concept lookup. The tools generate several linguistic annotations and features, which are used to build the extraction patterns to be matched in the text and extract the information.

The initial step is preprocessing the document structure to convert it to a format that can be handled by the extraction system (e.g., removing tags and original mark up from an HTML document and convert it into a raw text).
The input documents are unstructured and contain natural language text. The information is extracted by creating a set of patterns to be matched to the natural language text. Because of the complexity of natural language, it is difficult to describe these patterns as simply word sequences. We need first to structure the input text, identifying its essential lexical and syntactic constituents. The documents chosen (domain documents) are written in natural language so there are several linguistic preprocessing steps that must take place in order to structure the input text and make the knowledge accessible.
We use ANNIE (a Nearly-New Information Extraction system from the General Architecture for Text Engineering (GATE) \cite{maynard2016natural,Bontcheva2004,Cunningham2011a,Cunningham2002,cunningham2011text}  to process the text to obtain various linguistic features, which we use to build the patterns to be matched in the text and extract the information. ANNIE is composed of a set of processing resources that form a pipeline. The processing resources run in sequence over the documents; each resource creates some type of annotation and its corresponding features.
Text preprocessing includes tokenization, sentence splitting, shallow lexico-syntactic analysis (Part of Speech tagging and morphological analysis) and named entity recognition.
First, the tokenizer segments the text into logical units called tokens. The tokenzier produces ``Token''and ``SpaceToken''annotations for each token, with features orthography (capitalization form of words), kind (word, number, punctuation, or symbol), length, and string (used to look for a specific word). Next, the sentence splitter splits text into sentences and creates ``Sentence''annotations and ``Split''annotations on the sentence delimiters. Subsequently, The POS tagger module performs lexical analysis and adds a category (part of speech category, e.g., noun (NN), verb (VB), or proper noun (PNN)) feature to each Token annotation. Finally, the morphological analyser generates a ``root'' (root of the word or lemma) feature on Token annotations. This feature is used by the gazetteer.
Gazetteer Lookup: Gazetteers are plain text files containing lists of names (e.g. of rivers, cities, and people). Each gazetteer has an index file listing all the lists, plus features of each list (major, minor, language, and Annotation); e.g., the features for country.lst list (location, county, Lookup). The gazetteer generates Lookup annotations with relevant features corresponding to the list entry list matched. Lookup annotations are used primarily for named entity recognition \cite{cunningham2005information}.
In addition to the default ANNIE gazetteer lists, we have created a set of domain-specific gazetteer lists. The lists contain specialized terms which help us with identifying concepts relevant to our domain. More details about domain-specific lists are presented in Section (6.6.1). Further processing and the creation of matching patterns depend on the annotations and features produced by the respective ANNIE modules.

\subsection{LUSA Domain Ontology Construction.} The ontology is described as a formal specification of domain knowledge\cite{gruber1995toward}. In this context, the structure and type of the knowledge to be extracted from the selected documents is defined by the ontology. The criteria ontology encodes the categories of terms describing criteria (factors input to the suitability analysis model), their properties and the relationships that may exist among them, for which the selected documents can be searched. The ontology is provided as an input to the extraction system. The ontology guides the extraction process, providing the structure and the semantics of the knowledge to be extracted. The ontology is also populated with the extracted information.

Ontology represents the backbone of an OBIE system. An ontology is used to capture and represent domain knowledge. IE systems are, to a varying extent, domain related. One of the requirements of IE is that the type of content to be extracted must be defined a priori. In the context of information extraction, ontologies specify the structure and type of information to be extracted. The challenge is to use an ontology that is adequate for guiding information extraction from domain-related text.
Thus, in order to specify the information to be extracted, we defined an ontology to represent the knowledge of our domain and use this ontology to guide the information extraction process.
LUSA ontology was built from scratch because none of the existing ontologies covers the concepts in our application domain. The LUSA criteria ontology was explicitly developed to extract land use suitability criteria from documents related to Saskatchewan. The logic driving this approach was to get a basic prototype land use criteria domain-specific, ontology-driven information extraction system developed (built) and working at a satisfactory level of performance. However, because the terminology, as catalogued in the gazetteer lists that were developed, and the criteria classes in the ontology are, to a great degree, common across most or all documents in this domain, and because many of the extraction rules can potentially be applied to these other documents, the LUSA criteria ontology can subsequently be built upon incrementally to extend its capabilities to other documents in this domain. Recall that re-use is one of the primary factors favoring the use of ontologies. With this in mind, although documents for a specific area were the focus, other land use suitability documents were consulted during system development in order to introduce, from the outset, at least some degree of generality to the information extraction approach.

\begin{figure}
\centering
 \includegraphics[width=9.5cm ]{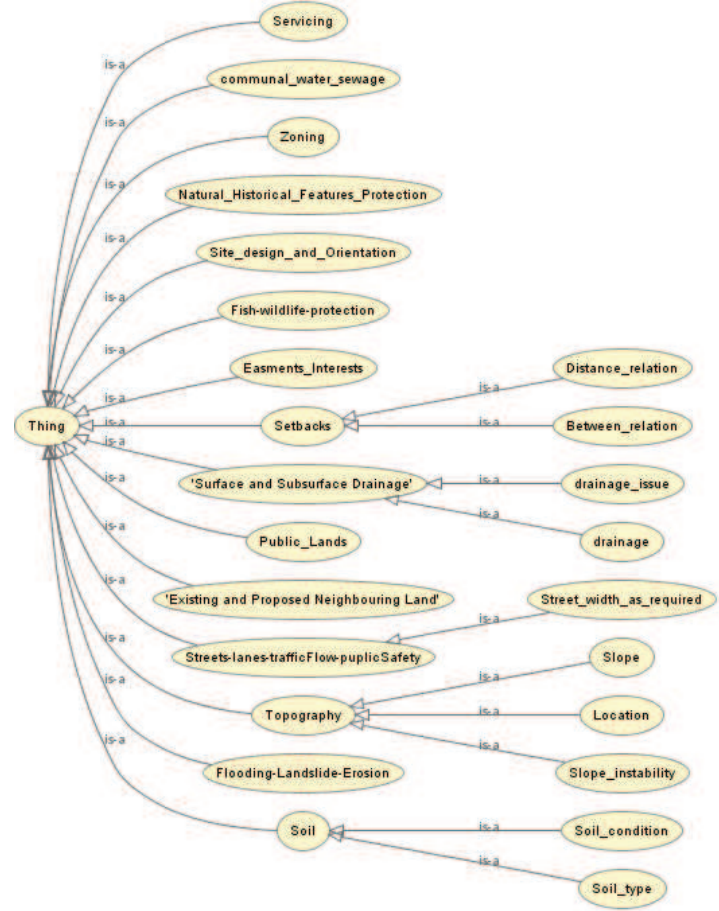}
\caption{LUSA Ontology Classes Hierarchy }
\label{fig62}
\end{figure}

From the analysis of domain-specific documents, we identified a list of factors that are considered by the city planners (CPB ''“ Community Planning Branch) when assessing land suitability for development. These criteria (factors) represent the biophysical, social and economic factors that may be used in the construction of land use suitability maps that support the process of evaluating the suitability of a particular area of land for a particular kind of use.
The ontology provides a model for the possible concepts (relations and objects) that might occur in the text. For example, the topography concept is modeled as a class and distance is modeled as a property for the class Setbacks. The semantic relations and hierarchy defined by the ontology provide an important support for natural language processing.
The process of developing the ontology is iterative and evolves with the analysis of domain text. Our first task was to identify the relevant concepts to be included in the ontology. LUSA ontology is organized into a hierarchy of sixteen main categories (as shown in Figure \ref{fig62}). The parent categories may have one or more sub-categories, which are specializations of the parent ones. For example, all the relations in the Setbacks category will belong to one of its subcategories (quantitative distance or qualitative distance).
The main categories of criteria are: Topography, Soils, Surface and sub-surface drainage, Potential for flooding and other hazards, Easements (or Interests), Land use (uses) in the vicinity, Streets, lanes, traffic flow and public safety, Site design and orientation, The protection of fish and wildlife habitats, The protection of significant natural and historical features, Setbacks, and Public Lands. In addition to defining the main categories and subcategories , for each concept, we identified the set of attributes that characterize it and its relations (if any) with other concepts; this is needed to make the structure of the body of knowledge explicit and to facilitate information access. For example, the quantitative distance relation in the setbacks category is characterized by the following properties: type of spatial relation (e.g., within, less than, or greater than), distance (i.e., a number that indicates a distance from an object), and object (e.g., landfill, water body, or intensive agriculture operation).
In our OBIE system, the ontology is not only used to guide the extraction process, the output is also stored in the ontology (ontology population).

\subsection{Ontology-Based Semantic Annotation.} In this step the domain ontology is coupled with a set of domain-specific gazetteer lists and pattern matching rules. Gazetteer lists contain names of instances of domain concepts (instances of classes and property values).  The documents are searched for instances of classes and property values defined in the ontology and instances of these concepts in the text are annotated with respect to classes and properties in the ontology.  A set of grammar rules are used to annotate and extract more complex patterns and structures than can be done by gazetteer lists. The grammar rules check if instances found in the text belong to a class  or a in the ontology and if so, they link the recognised instance to that same class and add ontology and class features to the annotations.

Semantic annotation \cite{cunningham2009developing,cimiano2005gimme,uren2006semantic,vargas2001knowledge} is the process of attaching metadata to selected parts of text to assist automatic interpretation of the meaning covered by the text. The task of the OBIE application is to identify instances in the text belonging to concepts in the ontology.
In ontology-based semantic annotation, the text is searched for instances of concepts (classes, relations and property values) defined in the ontology. For example, in the domain of land use suitability analysis, the input documents are searched for instances of classes such as Setback (distance, from what) and Topography (e.g., soil type, soil condition, and slope instability). The matched instances in the text are annotated with the relevant ontology concepts. Thus, instances in the text are linked to their ontological information (classes, relations and properties) via semantic annotation.
In our system, the ontology is coupled with domain-specific gazetteer lists (the gazetteers integral to ANNIE are very generic) and extraction rules to augment and enrich the semantic annotation process. Extraction rules are used to annotate and extract more complex patterns and structures than can be done by gazetteer list matching. The grammar rules operate on annotations generated by the linguistic processing modules and the Lookup annotations created by the gazetteer, including those created for the domain-specific lists, and combine them into more complex structures.

\subsubsection{Domain-Specific Gazetteer Lists}

The LUSA ontology concepts and their instances are included in the domain-specific gazetteer lists. The gazetteer lists contain specialized terms (instances of key concepts, for example, criteria factor, slope instability condition, soil type or drainage type, of the ontology concepts (i.e., classes, properties, or relations)). The ANNIE gazetteer \cite{cunningham2009developing} comes with a set of generic gazetteer lists used for named entity recognition, such as names of countries, cities, currencies and date. However, the ANNIE default gazetteer lists do not cover concepts in the domain of land use suitability assessment (or other specialized domains). Therefore, in addition to the generic default lists, we created a set of domain-specific gazetteer lists - one list for each class of criteria in the ontology (and for some subclasses and properties) e.g., landforms, slope, spatial relations. As previously mentioned, although we are using documents related to Saskatchewan, these terms were identified from a wider selection of documents in order to broaden the coverage of the gazetteer lists and the extraction system.
The terms in the domain-specific lists are used to identify occurrences of those terms in the text. When the gazetteer is run over the input text, a Look up annotation will be created for any text (word or phrase) that matches a list entry, and the features (e.g., major, minor, and annotation type) associated with the matched list will be added to the Look up annotation. For example, if an entry from the ``slope\_condition.lst''gazetteer list matches some text in a document, the gazetteer processing resource creates a Look up annotation and assign ``soil condition''to the majorType feature of the annotation for that text. Figure \ref{fig63} below, shows an example of the domain-specific ``soil\_condition.lst''list while Figure \ref{fig64} shows the annotations created after the gazetteer has processed the text and identified instances of domain concepts (e.g., spatial relation, soil condition, soil type, and drainage issue). A Lookup annotation is created for each term identified in text, and the major type of the matched list is assigned to the majorType feature of the annotation.

\begin{figure}
\centering
\includegraphics[width=9.5cm ]{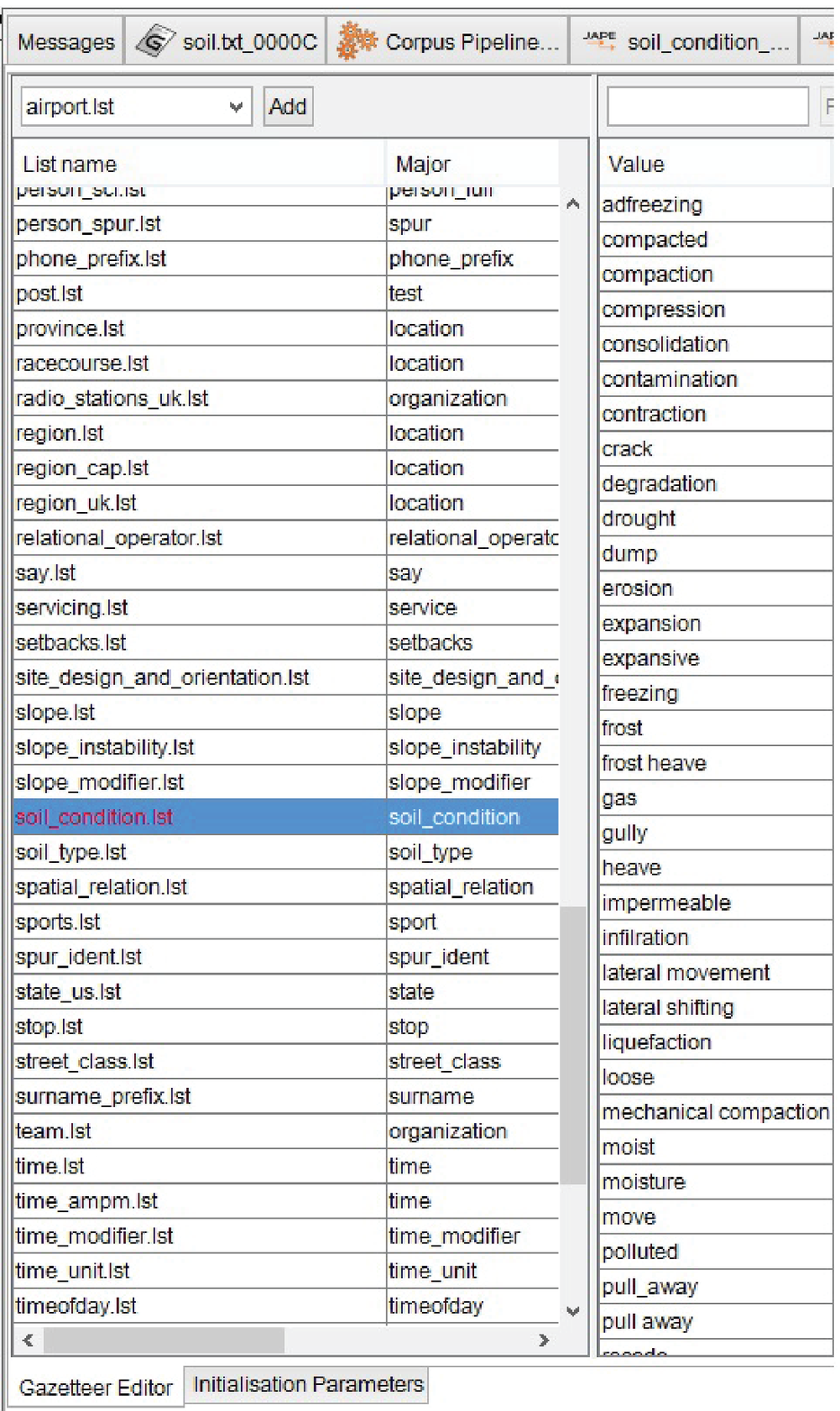}
\caption{A portion of the soil\_condition.lst list  }
\label{fig63}
\end{figure}

The result of this process is the identification and annotation of instances of domain concepts in the text. The annotations produced by the gazetteer lists are combined with the annotations produced by the linguistic processing modules to form extraction patterns used by the rules.

\begin{figure}
\centering
\includegraphics[width=9.5cm ]{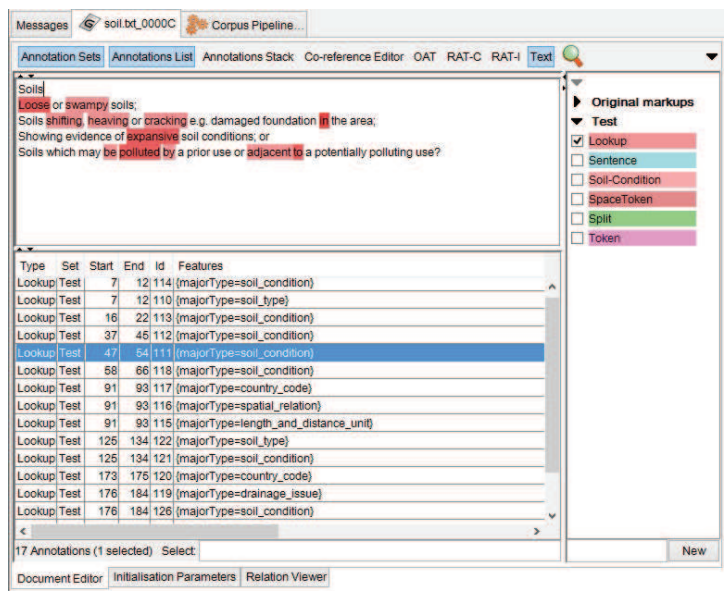}
\caption{Text annotated with entries from soil\_condition.lst list}
\label{fig64}
\end{figure}

\subsubsection{Pattern-Matching Extraction Rules}

\begin{figure}
\centering
\includegraphics[width=9.5cm ]{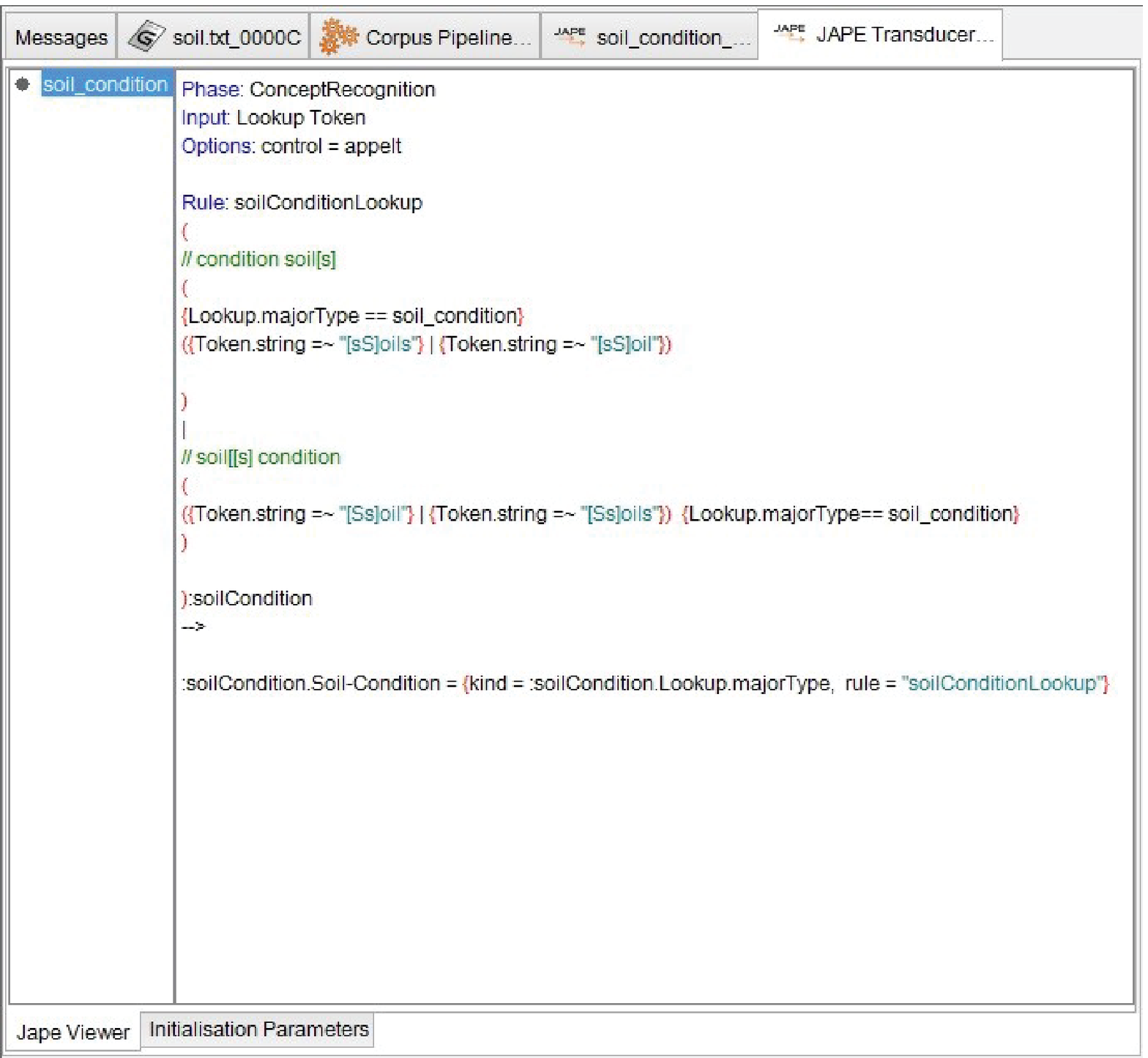}
\caption{An extraction rule that matches patterns where the entry from the soil\_condition.lst list appearing in the text is preceded by or followed by the word ``˜soil'' or ``soils''}
\label{fig65}
\end{figure}

Gazetteers can be used to find terms that suggest entities. However, the entries can often be ambiguous (e.g., some words or terms have several different meanings depending on the context). Figure \ref{fig65}, below, shows an example of a rule for checking if any word in the text matches any of the entries in the ``soil\_condition.lst''list. Note that although the rule is simple, it resolves ambiguity as it only matches patterns where the entry from the list appearing in the text is preceded by or followed by the word ``soil''or ``soils''.

Hand-crafted extraction rules or pattern-matching rules capture certain patterns in the text. The patterns are encoded as regular expressions in a language called JAPE \cite{cunningham2009developing}. JAPE is a Java Annotation Patterns Engine. JAPE provides finite state transduction over annotations based on regular expressions. Extraction rules combine annotations and features created by the gazetteer processing resource and linguistic processing resources (e.g., tokenizer, morpher and POS tagger) to further identify patterns/entities from the text.
These patterns can help combine different annotations involving a number of terms (phrase) that together provide information about a concept or a relation involving the concept. For example, to extract the information concerning the setback quantitative distance (e.g., within 500 meters of a water body), we developed a pattern-matching rule that combined major Lookup features created by domain-specific lists (e.g., spatial\_relation, distance\_unit and object) with linguistic annotation (such as preposition and determinant).
Our extraction system follows a knowledge engineering approach \cite{embley2004toward}, thus it relies on hand-crafted extraction rules. We have defined different sets of rules for extracting concepts, instances and property values and for populating the ontology.

{\bf 1. Grammar rules for the recognition of domain-specific concepts}

\begin{figure}
\centering
\includegraphics[width=9.5cm ]{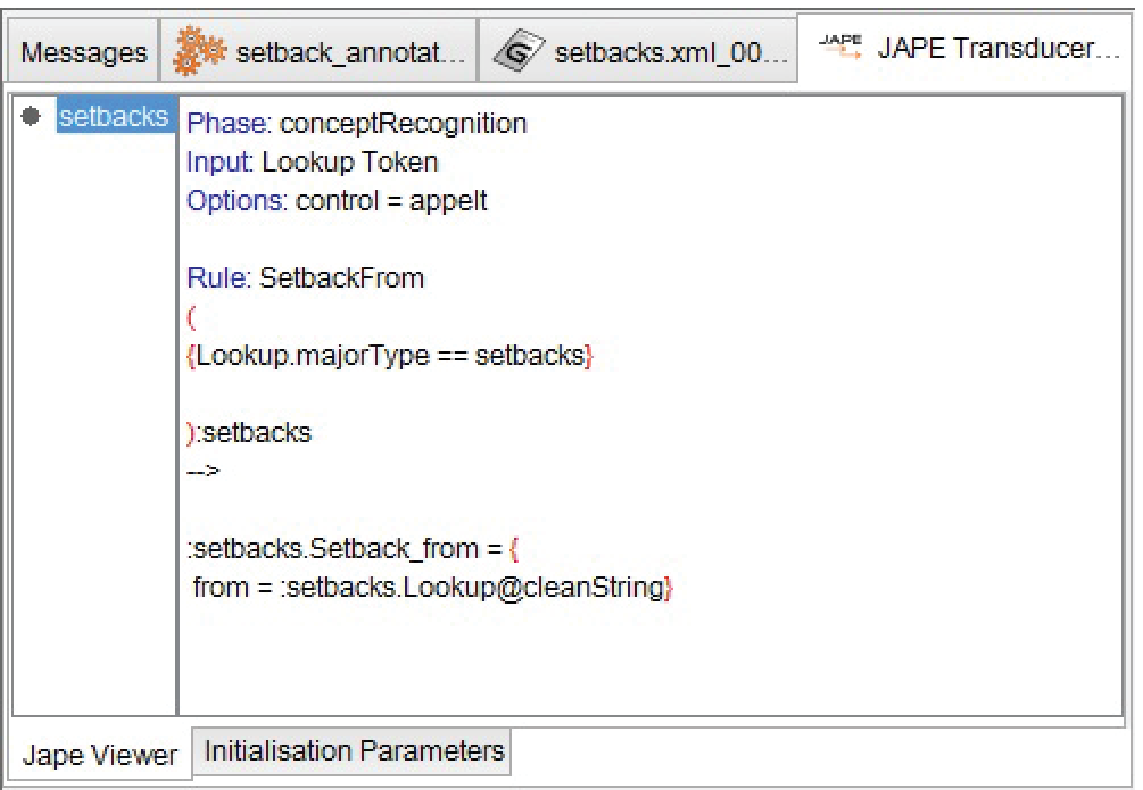}
\caption{A rule for identifying in the text terms (values) for 'setback\_from' property of the Setback class}
\label{fig66}
\end{figure}

\begin{figure}
\centering
\includegraphics[width=9.5cm ]{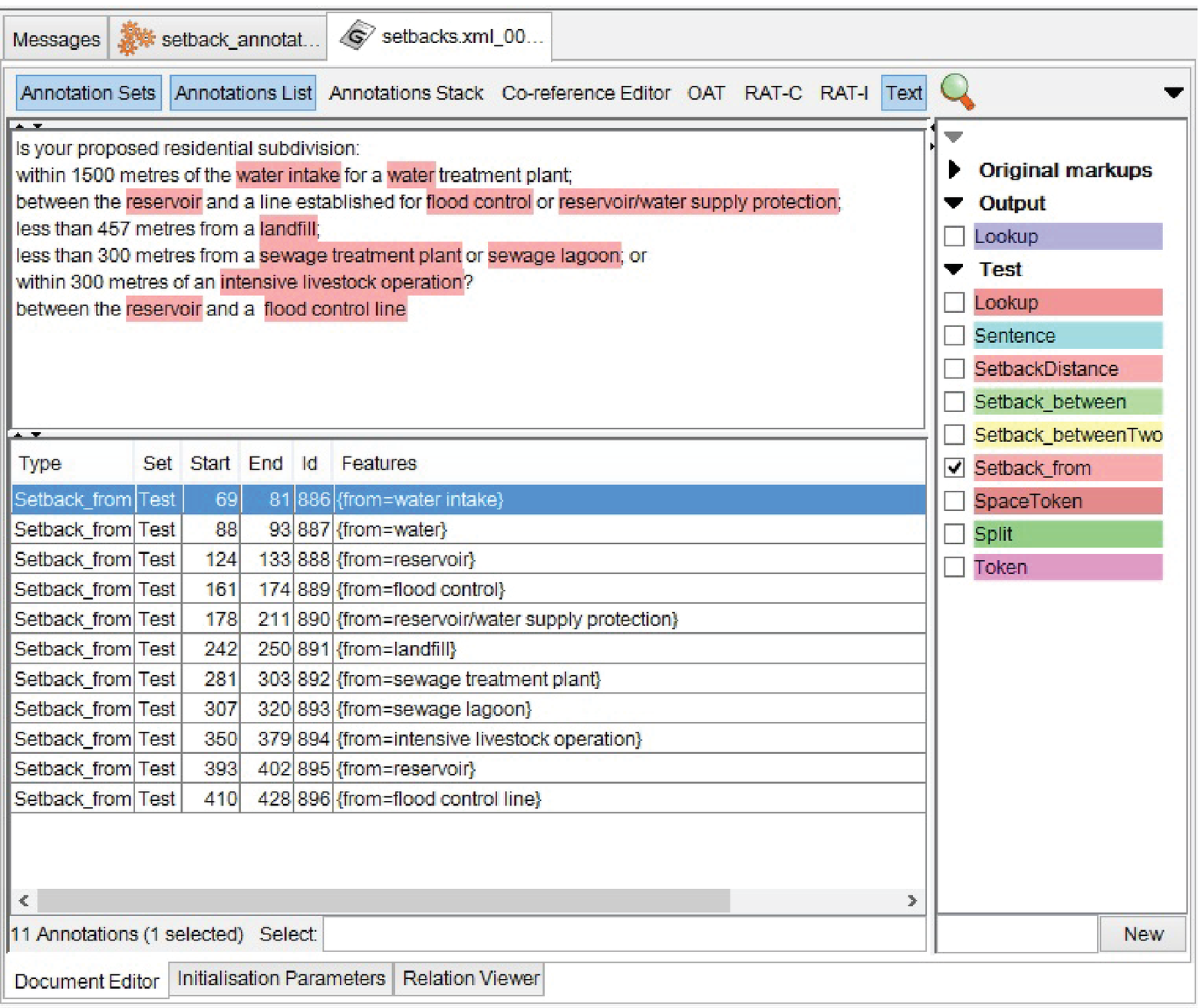}
\caption{ Identification and labeling of terms for the setback concept  }
\label{fig67}
\end{figure}

This type of rules are used for domain concept recognition and annotating them with the corresponding ontology concept. The grammar rule checks if a string in the text matches any of the entries in the gazetteer list associated to an ontology concept (class, property or relation). These rules mainly make use of Look up annotations generated by domain-specific gazetteer lists. Figure \ref{fig66} shows a rule for recognizing instances of the setback concept in the text, while Figure \ref{fig67} shows the annotation of recognized concepts in the text.

\begin{figure}
\centering
\includegraphics[width=9.5cm ]{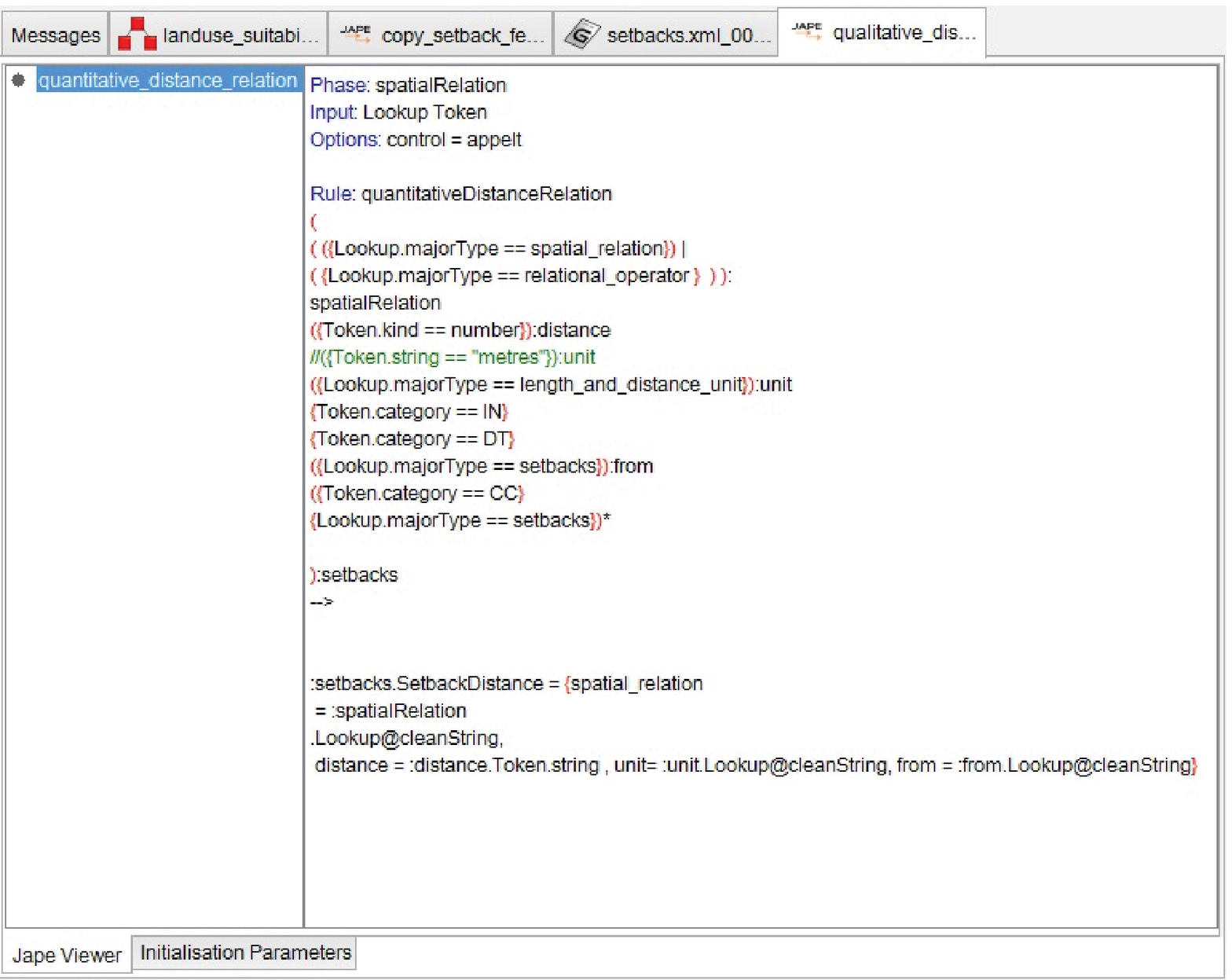}
\caption{A rule for identifying a complex pattern, representing a concept that has sub-concepts and properties}
\label{fig68}
\end{figure}

{\bf 2. Grammar rules for complex pattern matching}
Not all concepts can be recognized directly from gazetteer lists or with simple rules. Some entities require more complex rules based on contextual information. In addition,

the pattern to be matched may be not just a simple concept, but rather a concept that has several sub-concepts or properties or both. Figure \ref{fig68} is an example for a rule that matches a pattern for a complex concept.

\begin{figure}
\centering
\includegraphics[width=9.5cm ]{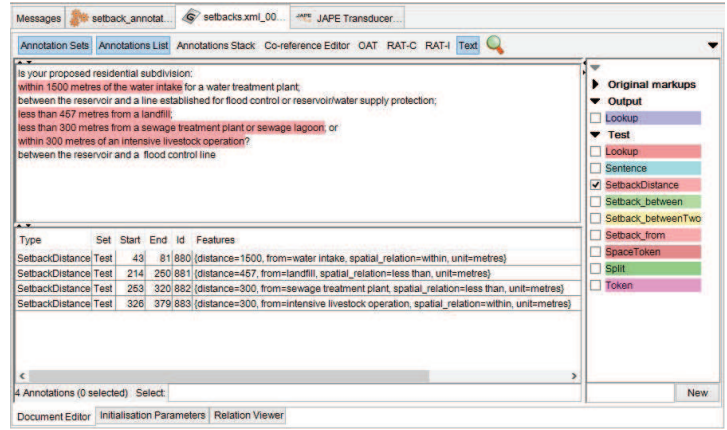}
\caption{ Annotation of a complex concept }
\label{fig69}
\end{figure}

\subsection{Ontology Population.} This component generates new instances in the ontology from the annotated text. Mentions of instances in the text are linked to instances of concepts in the ontology.
    \begin{itemize}
    \item Output Representation: The output of the system is presented as documents annotated with and linked to concepts in the ontology and the ontology populated with instances generated from the extracted information. The information in the populated ontology may be exported to a knowledge base, a database, an XML document or a text file for use or further analysis.
    \end{itemize}

The extracted information (the criteria and their values) are then integrated into the MCDM model applied for constructing the residential development suitability maps for the City of Regina.

\begin{figure}
\centering
\includegraphics[width=9.5cm ]{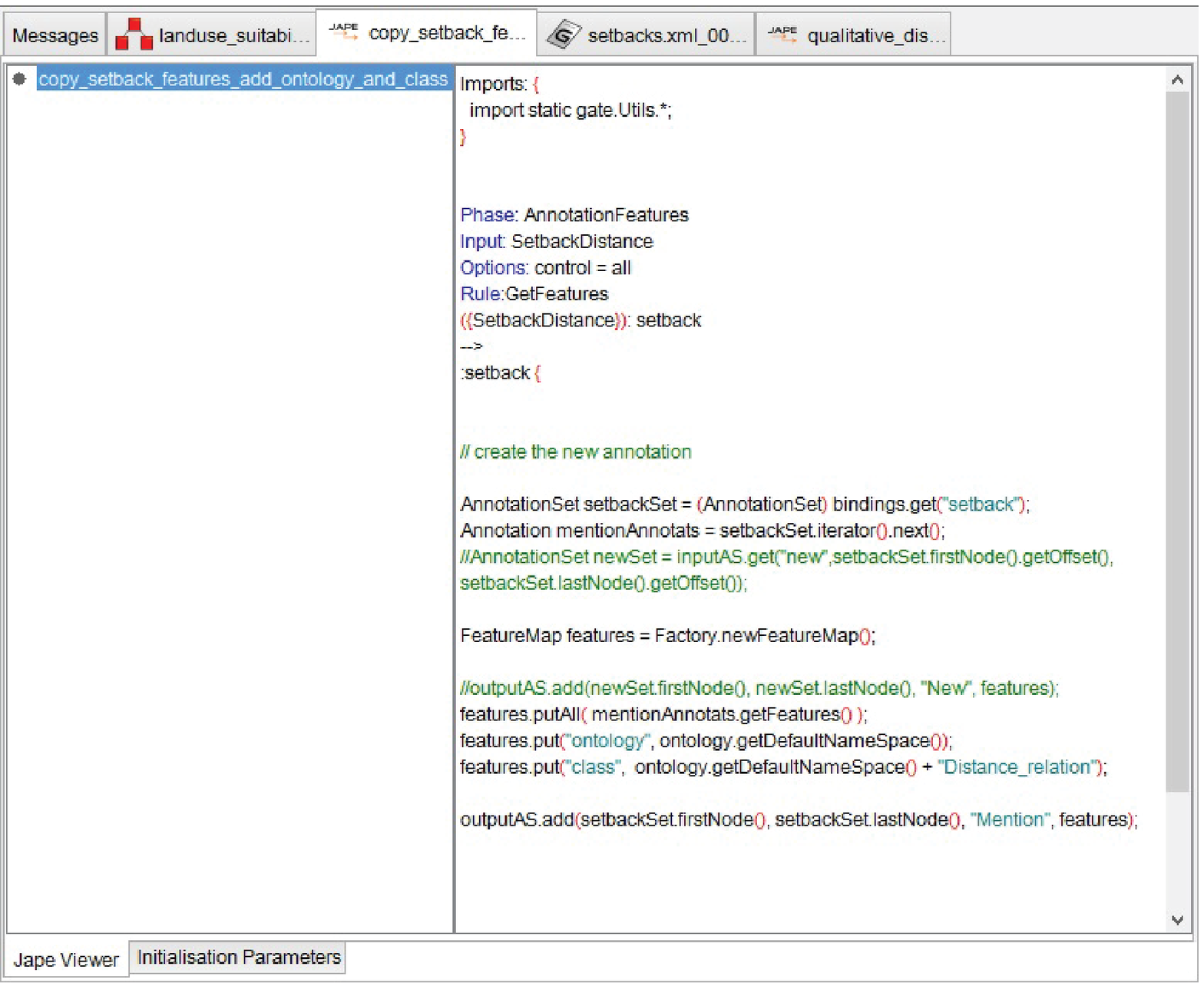}
\caption{  Linking Mention annotations to relevant ontology concepts}
\label{fig610}
\end{figure}

Ontology population is the process of adding instances of domain concepts to the ontology\cite{cunningham2011text,maynard2008nlp}. A concept instance is the mention of the concept in the domain corpus text; i.e., the occurrence of a concept as a term or a phrase in the text (e.g., the occurrence of the term ``river''as an instance of the class Drainage or the phrase ``100 m from a water body'' as an instance of the sub-class ``Quantitative\_Distance''of the Setback class or the term ``creeping''as an instance of the slope\_instability \_condition property of the class Topography.
The OBIE system is responsible for locating instances of domain concepts in text corpus and populating the ontology with those instances. Semantic annotation involves instance recognition and the annotation of instances in the text with relevant concepts from the ontology. After the documents have been semantically annotated, new instances are generated for the annotations. Finally, the new instances are mapped and added to the appropriate concepts in the existing ontology.
Annotations are linked to the ontology entity by having the URI of the entity in the ``inst''feature of the annotation. For example, the Setback annotation is linked to Setback class in the ontology by assigning the class name to the ``class''feature of the annotation. Semantic annotation facilitates the mapping of concept instances in the text to the relevant concepts in the ontology.
The annotation can have a class and ontology features. The URL and URI of the relevant ontology and relevant class are then assigned as values to the corresponding annotation features.
In addition to the pattern matching rules, we have built a set of rules for linking mentions of concepts in the text to their ontological information, generating instances of mentions, mapping them to the right concepts in the ontology and populating the ontology.
Our algorithm for populating the ontology with instances of domain concepts mentioned the text corpus is described below:

\begin{enumerate}
 \item Retrieving the annotation for the specific domain concept and identifying it as a ``Mention''. ``Mention'' annotations indicate that entities belonging to concepts in the ontology are mentioned in the input text.
 \item Linking ``Mention'' annotation to the ontology concepts by adding the relevant class name and ontology name to the annotation. This can simply be done by adding a feature ``class''and a feature ``ontology'' to the annotation and assigning the names of the relevant class and ontology as values to the corresponding annotation feature. For example, to relate the Setback annotation to its appropriate ontological class, we specify Setback as a value for the ``class'' feature of the annotation. Figure \ref{fig610} shows the rule for linking Mention annotations of Setback class to the Setback class in the ontology.
\item  Generating a new ontology instance for every Mention annotation.
\item Mapping and adding the instances to the right concepts in the ontology.
\end{enumerate}

\subsection{Output Presentation}

\begin{figure}
\centering
\includegraphics[angle=270,width=11cm ]{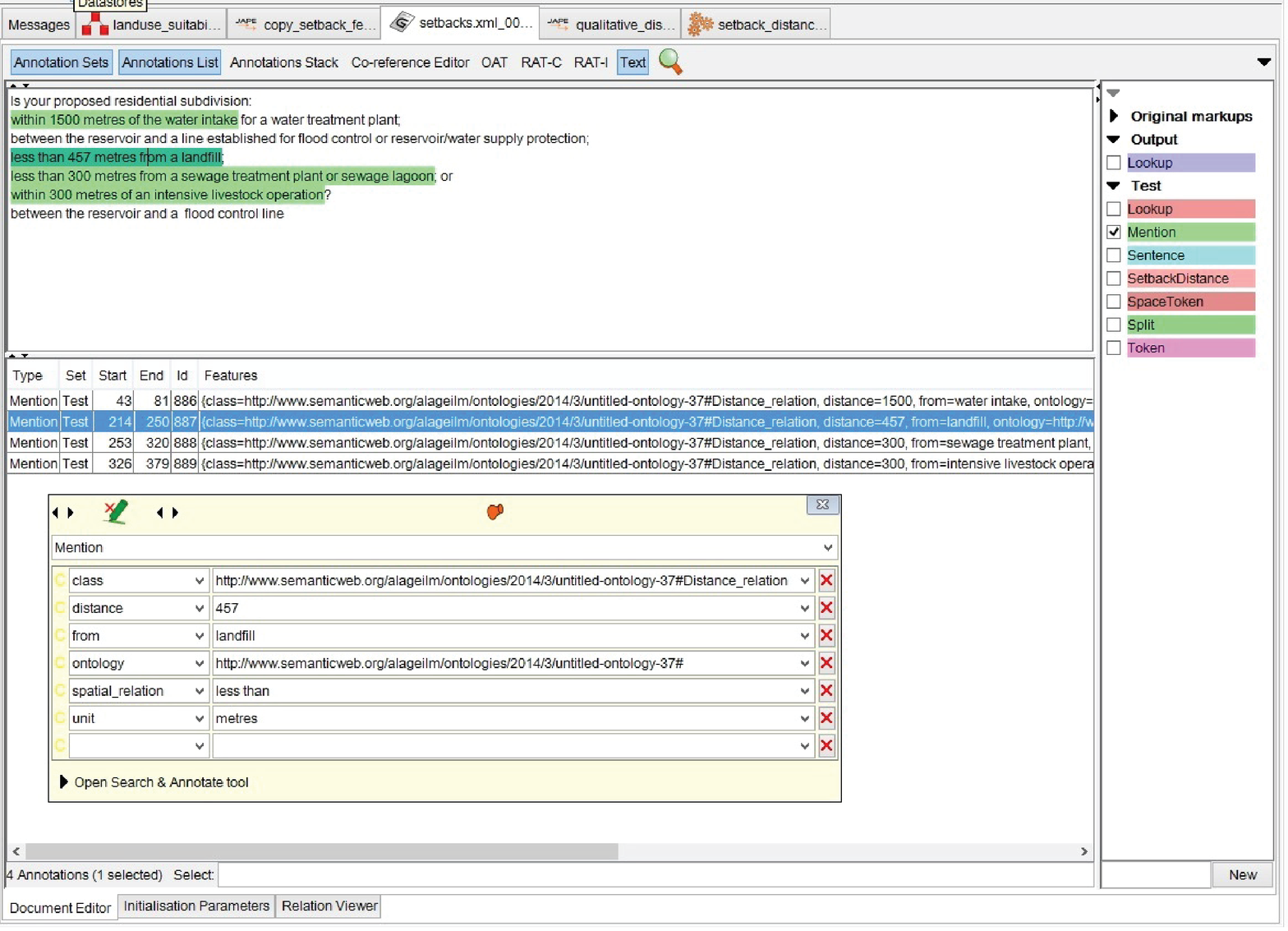}
\caption{Mentions of the Distance relation in the text annotated and linked to their ontology concepts}
\label{fig611}
\end{figure}

The output of the LUSA OBIE is presented as:

\begin{itemize}
\item Documents annotated and linked to concepts in the ontology. Figure \ref{fig611} shows an example of mentions in the text annotated and linked to their ontology concepts.
\item LUSA ontology populated with instances generated from the extracted information. The ontology is populated with instances and data property values. Figure \ref{fig612} below shows an example of the Setback class populated with instances generated from annotated text.

\item The information in the populated ontology may be exported to a knowledge base, a database, an XML document or a text file for use or further analysis.
\end{itemize}

\begin{figure}
\centering
\includegraphics[ angle=270,width=11cm ]{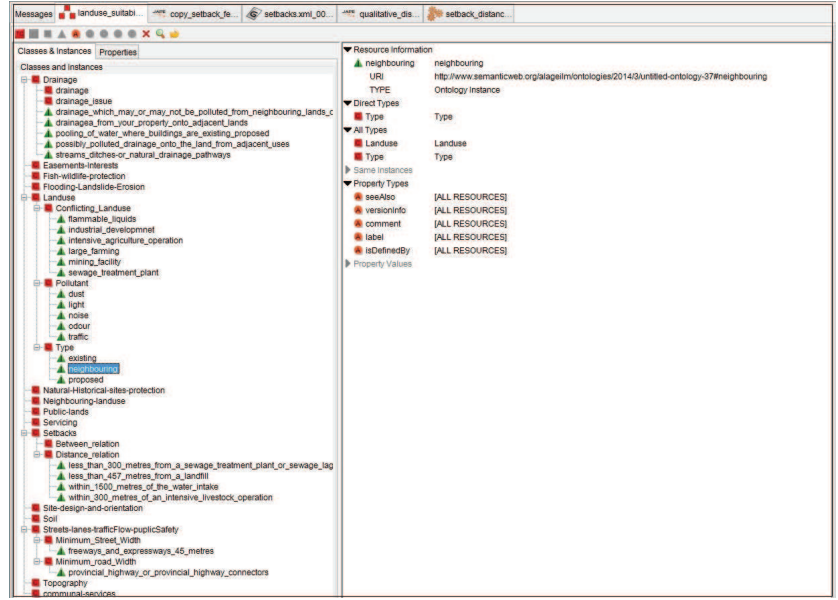}
\caption{ Populated Ontology }
\label{fig612}
\end{figure}

\section{GIS-based Multi-Criteria Decision Making (MCDM)}
\begin{figure*}
\centering
\includegraphics[width=10cm ]{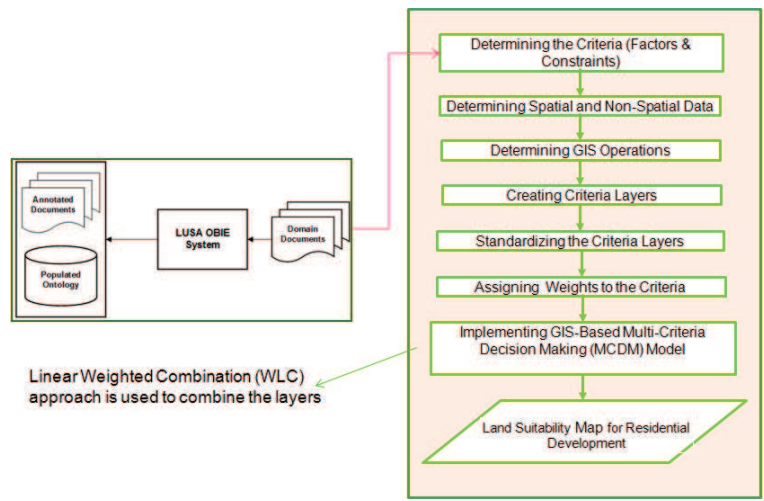}
\caption{Proposed LUSA OBIE and GIS-based multi-criteria evaluation for residential suitability}\label{f2}
\end{figure*}

GIS-based Multi-Criteria Decision Making (MCDM) analysis (also known as Multi-Criteria Evaluation (MCE)) \cite{eastman1998multi} was applied in this work. Multi-criteria analysis combines various criteria into a single evaluation index that indicates the relative suitability of different locations for a specified use, such as residential development.  Multi-criteria evaluation is determined through a method known as Weighted Linear Combination (WLC) \cite{jiang2000application}. Using WLC, the continuous criteria (factors) are standardized to a common numeric range, a weight is applied to each factor and then the weighted factors are combined to yield a weighted average.  The result is a continuous map of suitability that can be masked by the Boolean constraints to produce the final suitability map.
Figure \ref{f2} summarizes the steps included in a GIS-based multi-criteria evaluation procedure for land use suitability analysis. 

\section{{Case study: the city of Regina}}
\subsection{{ Study Area }}
\begin{figure}
\centering
\includegraphics[width=9.5cm]{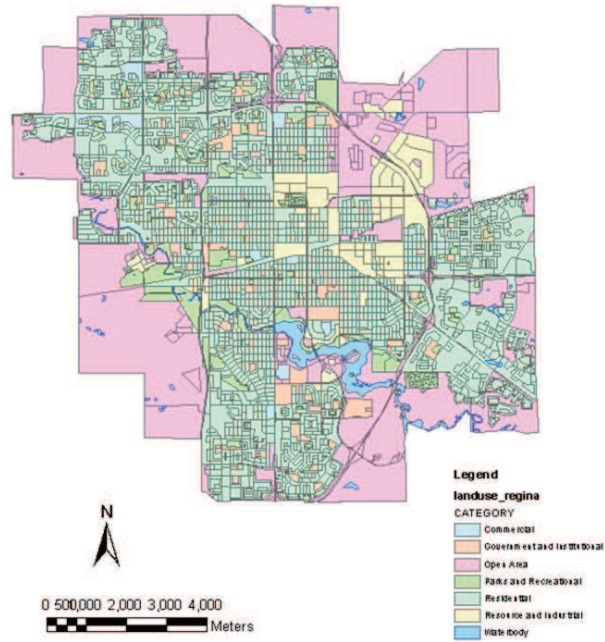}
\caption{ Land use map for the City of Regina}\label{f1}
\end{figure}

Regina is the capital city of the Canadian Province of Saskatchewan. It is located at   50''° 27' 0" N / 104''° 37' 0" W. Regina is the second-largest city in Saskatchewan and represents a cultural and commercial centre for the southern part of the province. Figure \ref{f1} shows a land use map of the study area. Regina is experiencing both economic and population growth. According to the Statistics Canada, 2016 Census of population, the population grew to 215,106 in 2016 compared to 193,100 in 2011 and to 179,282 in 2006 (Statistics Canada).

\subsection{ {Specification of the Criteria }}
The first step in a multi-criteria analysis is determining the criteria to be used. The criteria are of two types:
\begin{itemize}
\item Constraints: the regulations that limit the area available for development (those areas that are not suitable or not allowed for development under any circumstance \cite{eastman1999multi}; e.g., water bodies and already developed areas are restricted from development); constraints are thus Boolean.

\item  Factors: criteria that determine the relative suitability of the remaining areas for residential development; factors are continuous \cite{eastman1999multi}. For example, the type of existing land use is a factor that can increase or decrease the suitability of the land for development or, to a certain degree, areas close to major roads are preferable for development over areas that are distant from major roads.
\end{itemize}

Using our LUSA OBIE, we were able to identify a set of criteria pertaining to land use suitability analysis for residential development in Saskatchewan.
For the purpose of this study several criteria were selected.

Constraints include the following: new development cannot occur within 100 meters of water bodies, areas that are already developed, and water bodies and roads are not considered suitable for new development under any circumstances.

Factors are as follows.
\begin{itemize}
\item	current land use type: after eliminating areas that can be considered, remaining areas are rated according to their type; e.g., open areas are preferred to treed areas,
\item	distance to major roads, and
\item	distance to existing developed areas.
\end{itemize}

Other attractive, but non-essential, factors such as distance to schools and hospitals can be considered, but were not included in this analysis.

 \subsection{ {Selection of Data Layers }}
Following the selection of the criteria, the required image data necessary for the creation of factor and constraint layers (which will be combined to produce the final suitability map), were downloaded. The following datasets were used: land use map of Regina, road network, lakes and rivers, and wetlands. DMTI Spatial Inc. data layers were accessed via the Equinox website: \\
http://equinox.uwo.ca/EN/AdvancedSearch.asp.
\subsection{ {Creation of Factor and Constraint Images }}
\begin{figure}[h]
\centering
\includegraphics[width=5cm ]{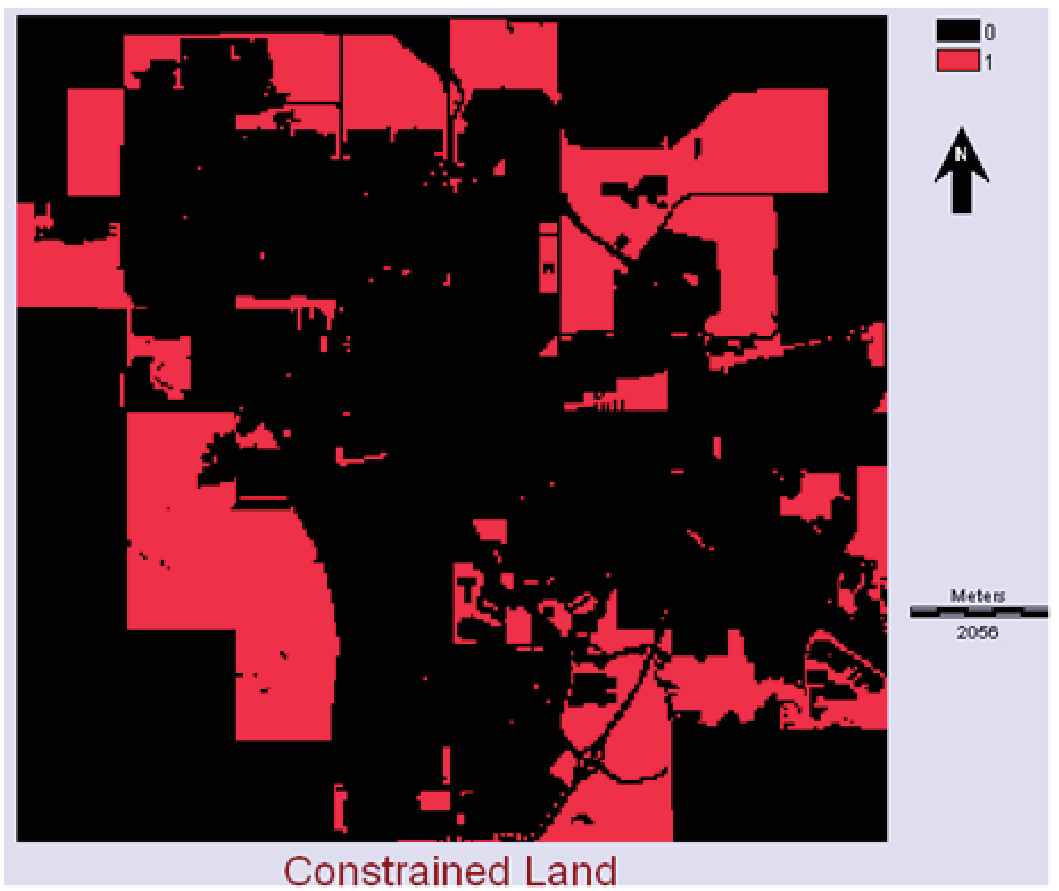}\includegraphics[width=5cm ]{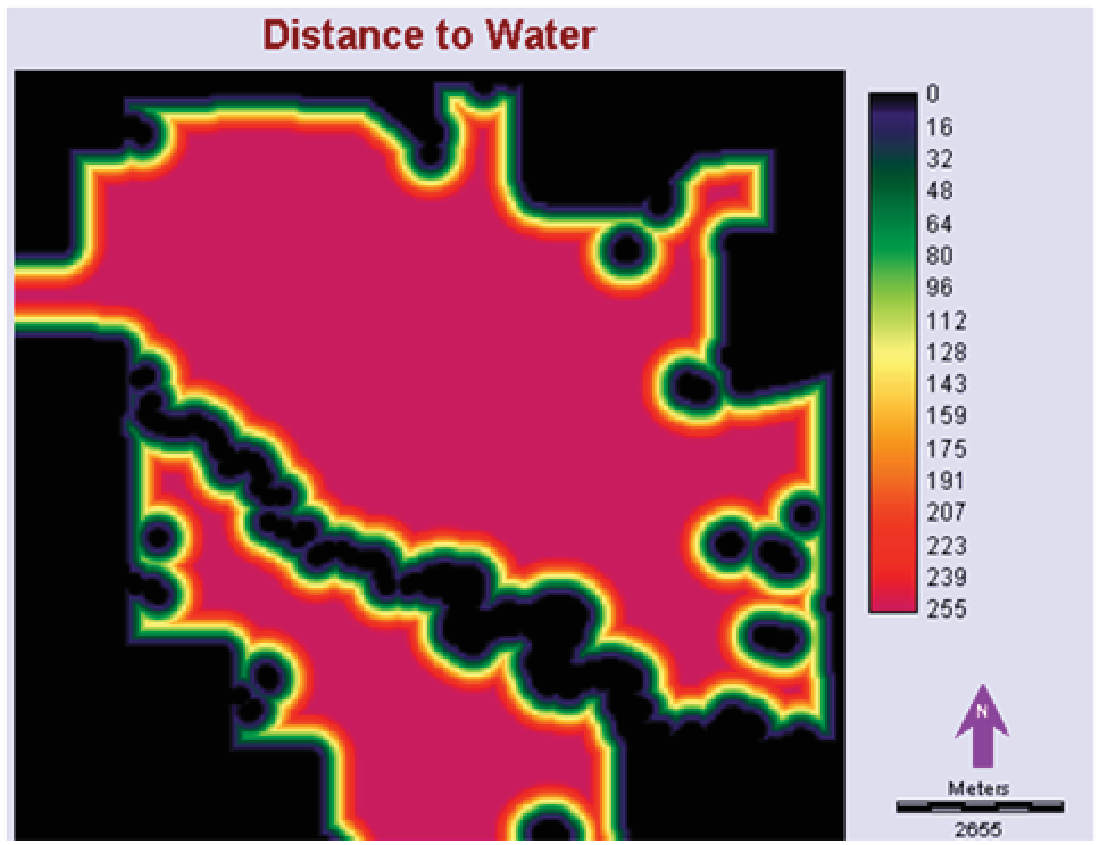}
\includegraphics[width=5cm ]{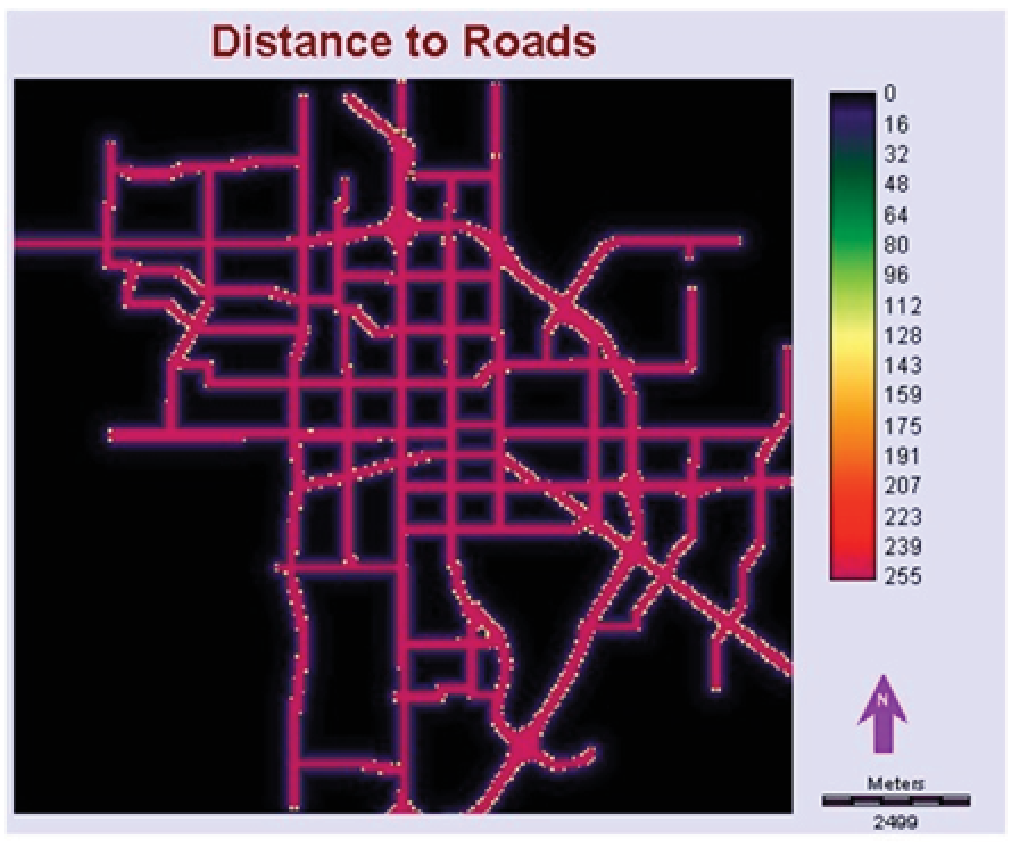}\includegraphics[width=5cm ]{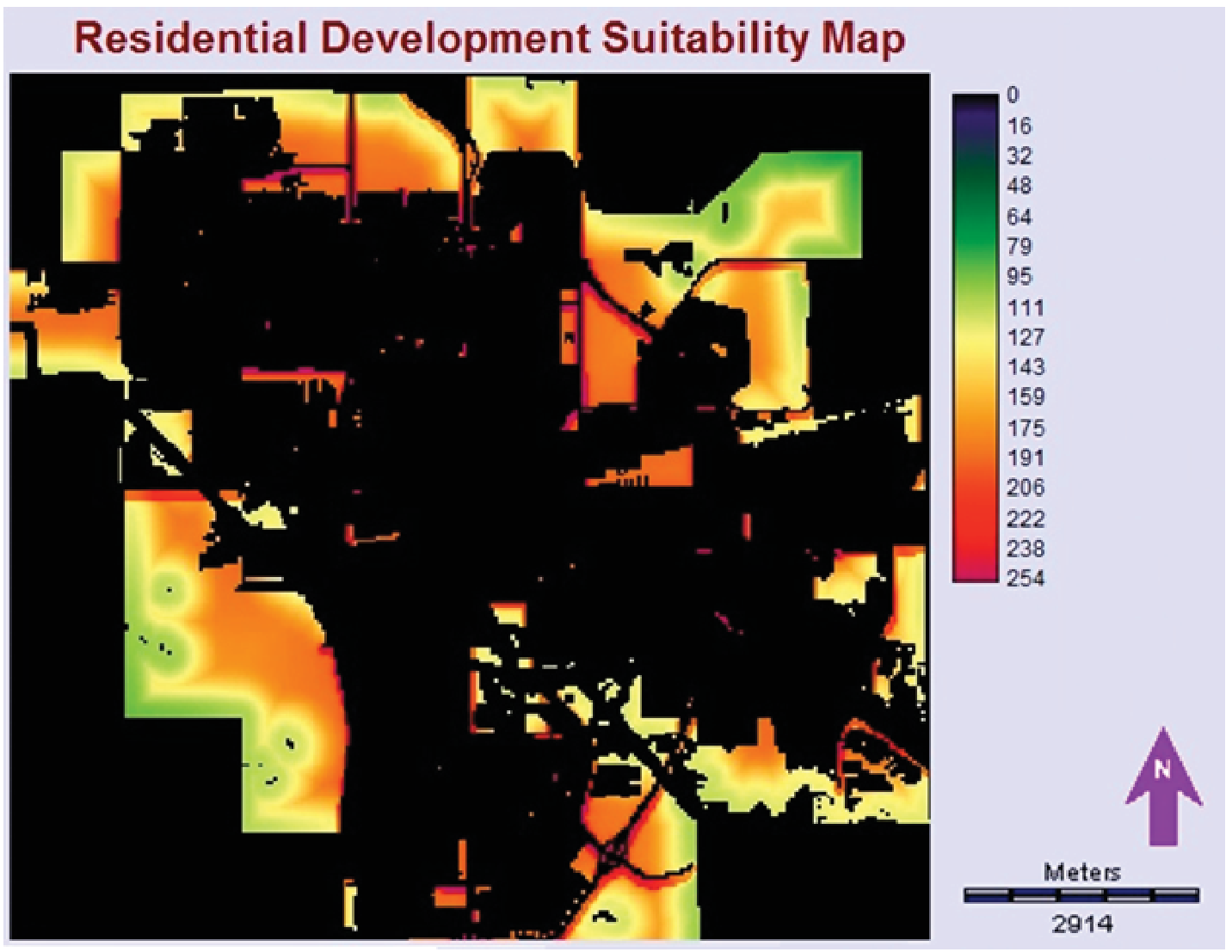}
\caption{Constrained Land (top left), Distance to Water (top right), Distance to Roads (bottom left) and Residential Development Suitability Map (bottom right)}\label{f4}
\end{figure}

A raster image was created for each constraint and factor.  The geospatial processes used to create the layers that represent the criteria applied in land use suitability analysis are reclassify, overlay, and distance.  The raster images are shown in Figure \ref{f4}.  The factor images created have different measurement units. To enable combining these factors, the images were standardized to a continuous scale from 0 to 255, where non-suitable areas are represented by 0 and the most suitable areas are represented by the value 255. Constraint images remain Boolean, where non-suitable areas are assigned the value 0 and suitable areas are assigned the value 1. For the purpose of this work, the criteria were assigned equal weights. The Weighted Linear Combination approach was used to combine the layers and produce the final suitability map (see bottmo right of Figure \ref{f4}) for residential development.



~ \newpage
\section{{Conclusion}}
Land use suitability maps can be useful to planners, developers, and environmentalists in their discussions and in making informed decisions on future, sustainable development.  These maps can also be an effective means of presenting land use information to the public.  The process of extracting the criteria and developing such maps also serves to identify information that should be important to the decision process but is not readily available:  this may thus serve to initiate or support efforts to obtain such data.
The extracted information can also be applied to assess the suitability of land for other types of development, such as agricultural, industrial or commercial, and to create the desired suitability maps. The resulting maps may then be integrated into a simulation model, such as cellular automata, to predict future growth in the City of Regina. The LUSA OBIE system assists in this process by automating the identification of the criteria and the data that must be obtained to carry out land use suitability analysis.

In the near future we are planning to expand the knowledge extraction rules in order to extract other forms of information such as tabular data, image data, and spatial and temporal data and relations. These latter will be expressed in the form of spatio-temporal constraints and preferences using the models and techniques we have developed in the past years \cite{ALANAZI2020103182,mouhoub2018learning,mouhoub2008SCC}.

\bibliographystyle{plain}
\bibliography{muniraLusa2021}







\end{document}